\documentclass[runningheads]{llncs}

\usepackage{eccv}

\usepackage{eccvabbrv}

\usepackage{graphicx}
\usepackage{booktabs}

\usepackage[accsupp]{axessibility}  

\usepackage{caption}
\usepackage{makecell}    
\usepackage[table]{xcolor}
\usepackage[most]{tcolorbox}
\usepackage{multirow}
\usepackage[normalem]{ulem}

\usepackage{xcolor}
\usepackage[breaklinks,colorlinks,allcolors=eccvblue]{hyperref}

\begin{document}

\title{Why Do Vision Language Models Struggle To Recognize Human Emotions?} 


\author{Madhav Agarwal,
Sotirios A. Tsaftaris,
Laura Sevilla-Lara, 
Steven McDonagh}

\authorrunning{Agarwal et al.}

\institute{The University of Edinburgh, UK
\\
\email{\{madhav.agarwal, s.tsaftaris, l.sevilla, s.mcdonagh\}@ed.ac.uk}}

\maketitle
\begin{abstract}
Understanding emotions is a fundamental ability for intelligent systems to be able to interact with humans. Vision-language models (VLMs) have made tremendous progress in the last few years for many visual tasks, potentially offering a promising solution for understanding emotions. However, it is surprising that even the most sophisticated contemporary VLMs struggle to recognize human emotions or to outperform even specialized vision-only classifiers.
In this paper we ask the question `Why do VLMs struggle to recognize human emotions?', and observe that the inherently continuous and dynamic task of facial expression recognition (DFER) exposes two critical VLM vulnerabilities.
First, emotion datasets are naturally long-tailed, and the web-scale data used to pre-train VLMs exacerbates this head-class bias, causing them to systematically collapse rare, under-represented 
emotions into common categories.
We propose alternative sampling strategies that prevent favoring common concepts. Second, temporal information is critical for understanding emotions. However, VLMs are unable to represent temporal information over dense frame sequences, as they are limited by context size and the number of tokens that can fit in memory, which poses a clear challenge for emotion recognition. 
We demonstrate that the sparse temporal sampling strategy used in VLMs is inherently misaligned with the fleeting nature of micro-expressions (0.25--0.5 seconds), which are often the most critical affective signal. 
As a diagnostic probe, 
we propose a multi-stage context enrichment strategy that utilizes the information from `in-between' frames by first converting them into natural language summaries. This enriched textual context is provided as input to the VLM alongside sparse keyframes, preventing attentional dilution from excessive visual data while preserving the emotional trajectory. 

\let\thefootnote\relax
\footnotetext{
    Project Page: 
    \href{https://madhav1ag.github.io/vlm-temporal-emotion-gap}{https://madhav1ag.github.io/vlm-temporal-emotion-gap}
}

\end{abstract}

\section{Introduction}
\label{sec:introduction}
\begin{figure*}[htbp]
  \centering
  \includegraphics[width=\linewidth]{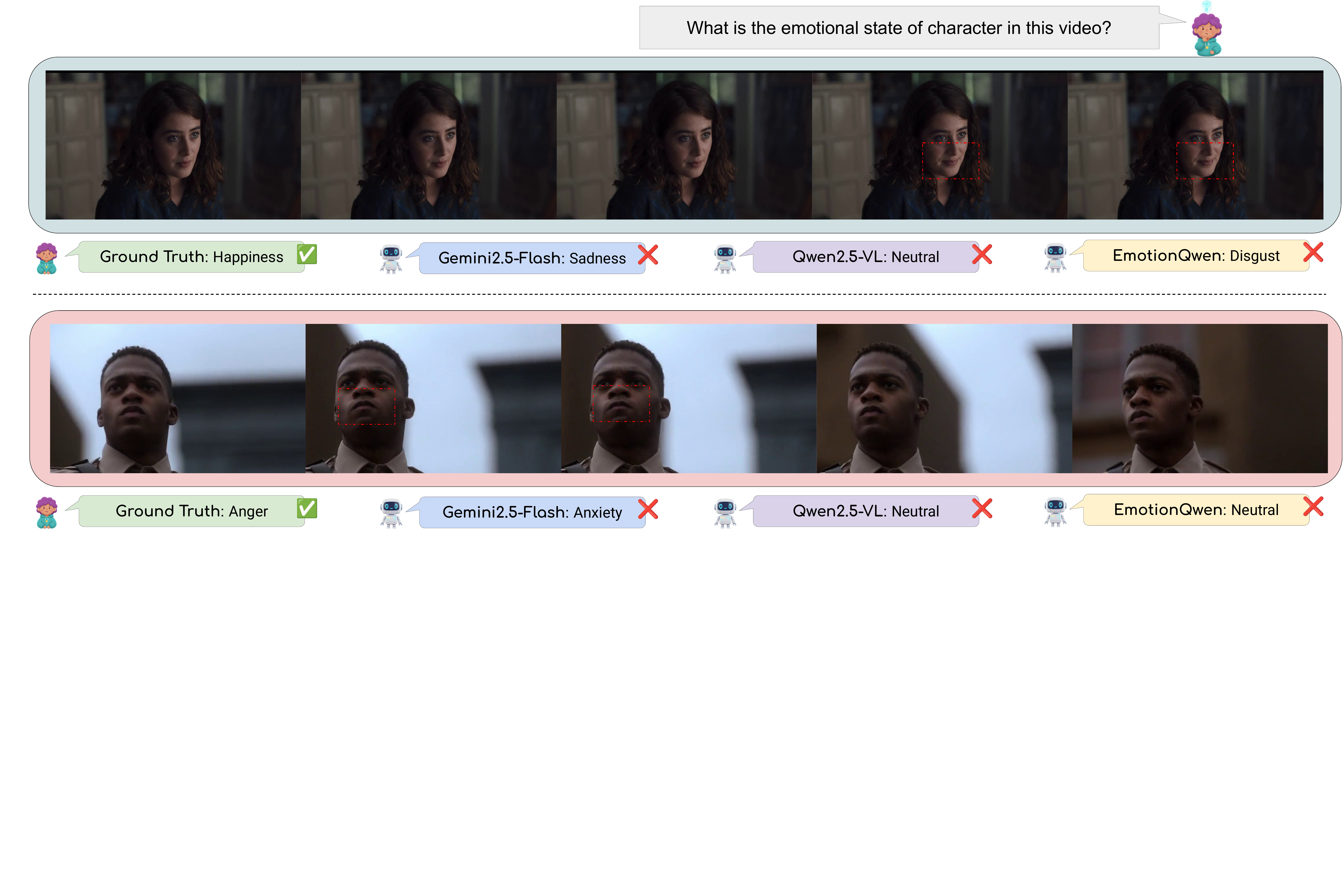}
  \vspace{-0.5cm}
  \caption{Qualitative Failure Analysis of SOTA VLMs on Video Emotion Recognition. VLMs like Gemini2.5-Flash, Qwen2.5-VL and EmotionQwen misclassify emotions by failing to capture subtle, temporal cues. Examples include overlooking a brief smile (top row, `Happiness' misclassified as `Sadness/Neutral/Disgust') or a tensed facial expression (bottom row, `Anger' misclassified as `Neutral/Anxiety')}
  \label{fig:motivation}
  \vspace{-0.5cm}
\end{figure*}

Comprehending human emotion is central to natural social interaction and a prerequisite for trustworthy, human‑centred AI. Emotion recognition infers a person's affective state from behavioral signals (e.g., facial expressions, voice, gesture), requiring the integration of subtle, dynamic patterns that unfold over time.
Humans excel at this by picking up on fleeting micro‑expressions (0.25--0.5 seconds)~\cite{haggard1966micromomentary,ekman1969nonverbal-mircoexpression,matsumoto2011evidence-microexpression}, modulate interpretations by context, and cope with a naturally long‑tailed distribution in which common states (\eg, neutral) dominate while rare but consequential emotions occur infrequently.

Equipping digital systems with comparable sensitivity would bring tangible benefits: from more supportive mental-health screening and safer conversational agents to richer assistive technologies for education and care.
Realising these gains requires models that respect the temporal nature of affect and perform robustly across both common and rare emotional states, aligning machine perception more closely with the way humans read and respond to one another.

Contemporary Vision–Language Models (VLMs) such as the Gemini-series ~\cite{comanici2025gemini,team2023gemini} and Qwen-series~\cite{Qwen2.5-VL,Qwen2-VL,Qwen-VL,bai2023qwen} can demonstrate strong performance when recognizing static objects and in grounding long textual descriptions. However, they do not yet exhibit robust emotion recognition abilities. Inferring human affective state from video demands temporal modeling to track how emotions evolve~\cite{jack2014dynamic-emotion} and the capturing of short‑lived micro‑expressions. In practice, contemporary VLMs often conflate closely related states (\eg, sadness versus disappointment). 
We hypothesize this gap arises because the task of dynamic emotion recognition inherently relies on overcoming two long-standing challenges that contemporary VLMs are uniquely vulnerable to: firstly, the inherently long‑tailed distribution of affective categories, which induces head‑class bias, whereby VLMs over‑predict common emotions and under‑recognize rare but consequential concepts. Secondly, inadequate temporal representation under limited context windows (token budgets), which promotes a `bag-of-frames' sampling strategy. This is fundamentally misaligned with the fleeting nature of micro-expressions and disrupts the temporal dynamics essential for accurate emotion classification.

We investigate the root causes of these VLM performance deficiencies.
We ask three questions: 1) How large is the performance gap between VLMs and vision-only classifiers in recognizing human emotions from video? 2) Why does this gap arise? and 3) How can it be mitigated? 

We adopt video‑based emotion classification as a probe task; it couples fine‑grained spatial sensitivity to subtle facial muscle movements with temporal fidelity and short‑lived micro‑expressions, making it a stringent benchmark for high-fidelity temporal modeling.
This setting also reflects the intrinsically long‑tailed distribution of affective categories. 
We quantify failures, linking them to two root causes: (i) head-class bias from long-tailed data and (ii) weak temporal modeling under fixed token budgets that promote sparse, order-agnostic frame sampling.
 
On our first issue, we highlight that VLMs trained on internet-scale data induce natural biases and inherit skewed concept and class‑frequency distributions~\cite{parashar2024neglected,mehrabi2021surveyofbias}. 
Since underlying training data specifications are often proprietary, we cannot directly quantify emotion class prevalence. 
However, we note that vision-language datasets are predominantly constructed by scraping web images and their associated HTML alt-texts. Because VLMs align visual features with these text embeddings during contrastive pre-training, the lexical frequency of an emotion concept in human-generated text directly dictates its representation density in the multimodal latent space. Building on this premise, we utilize historical lexical frequencies of emotion terms in Google Books Ngrams~\cite{lin2012googlengram,michel2011googlengram} as a representative proxy for this pre-training distribution bias.
This provides a high-level correlative indicator of conceptual prevalence 
in human-generated data, where we observe a pronounced long-tail. 
Class sparsity is inversely associated with per-class classification accuracy. Rarer emotions incur larger error rates in 
contemporary VLMs (see Sec.~\ref{subsec:long-tail-data-distribution}). The direct, empirical validation of this hypothesis, via controlled dataset balancing, is presented in Sec.~\ref{subsec:mitigate-long-tail}.

Our second key issue stems from observations that VLM abilities to retain and prioritise relevant information degrade as context length increases~\cite{liu-etal-2024-lost,wu2025visualhaystack}. 
This limitation is exacerbated in subtle video understanding tasks such as emotion recognition, where fine‑grained cues are diluted by redundant frames under a fixed token budget. 
To assess temporal understanding we vary frame‑sampling strategies and the effective frame rate, and probe order sensitivity by shuffling frames. 
We observe a quasi-bell-shaped relationship between performance and the number of frames available during inference: accuracy initially improves with denser evidence but declines once the token budget is saturated. Even at its peak, we observe that VLM performance remains below vision‑only classifier baselines. Moreover, we observe that accuracy is largely invariant to frame ordering, indicating weak temporal modelling and a reliance on order‑agnostic aggregation of per‑frame appearance. 

To mitigate these two fundamental failures, we propose corresponding `plug-and-play' solutions. To address the long-tail bias, we demonstrate the efficacy of a decoupled training strategy using a balanced dataset (Sec.~\ref{subsec:mitigate-long-tail}). More centrally, to solve the temporal bottleneck, we introduce a multi-stage context enrichment strategy. This inference-time pipeline addresses the core conflict between temporal density, needed for micro-expressions, and token budget limitations.
Instead of merely selecting from a sparse set of frames which risks information loss from induced gaps, our approach enriches the context. We firstly use a VLM to perform modality translation on the `in-between' frames, converting subtle motion cues into high-level natural language summaries.
This enriched textual context, which VLMs excel at processing, is then fed to the VLM alongside the original sparse keyframes to perform final classification, allowing the model to capture the fleeting dynamics it would otherwise have missed.
Crucially, our objective is not to engineer a task-specific model to marginally surpass current Dynamic Facial Expression Recognition (DFER) state-of-the-art benchmarks. Specialized models utilizing 3D-CNNs or spatial-temporal transformers currently dominate these metrics, but they lack the open-ended reasoning, zero-shot generalization, and conversational capabilities of VLMs. 
Our primary contribution is therefore not to chase these benchmarks but to provide a new understanding of VLM behavior for affective computing.
We systematically diagnose why generalized, massive-scale VLMs fail at affective tasks, and demonstrate that unlocking their temporal bottlenecks is a critical stepping stone toward achieving human-level proficiency in affective scene understanding.

Our main contributions can be summarised as:
\vspace{-0.24cm}
\begin{itemize}
    \item We provide empirical evidence that long‑tailed distributions, characteristic of internet‑scale data, correlate with per‑class accuracy, explaining underperformance in the emotion recognition task.
    \item We analyse temporal limitations of contemporary VLMs, using video‑based emotion classification as a probe to quantify performance and dissect root failure causes.
    \item As a secondary, diagnostic intervention, we propose a plug‑and‑play strategy that strengthens temporal modeling while keeping track of micro-expressions, within fixed token budget, yielding consistent improvements in performance on emotion classification tasks.
\end{itemize}
\vspace{-0.5cm}
\section{Related Work}
\label{sec:relatedwork}
\subsection{Vision-Language Models for Videos}
The rise of Large Language Models (LLMs)~\cite{touvron2023llama,bai2023qwen,cai2024internlm2,bi2024deepseek} has driven the use of vision encoders with language models to handle multi-modal tasks. 
Vision-Language Models~\cite{Qwen2.5-VL,achiam2023gpt4,team2023gemini,comanici2025gemini} have shown promising results in tasks like image captioning and visual question answering. 
These approaches are further extended to incorporate videos directly~\cite{alayrac2022flamingo,xu2025qwen2omni,maaz2024video-chatgpt,zhang2023video-llama}, by employing cross-modal attention and aligning video features with the LLM latent space. 
Although these models demonstrate promising early results on coarse-grained video understanding tasks~\cite{huang2024vtimellm,wang2023learningvat}, they exhibit limited capacity to attend to specific video segments and to perform temporal reasoning~\cite{liu-etal-2024-lost,arnab2025temporalcot,upadhyay2026timeblindness,ding2025dollmtime}. 

In particular, recent benchmarks highlight that VLMs often exploit spatial and textual biases rather than temporal reasoning~\cite{cores2024tvbench}, revealing a critical disparity between their observed performance and their underlying reasoning capabilities. 
The performance deficit in temporal reasoning stems from several foundational failures. 
Overly long context windows may introduce large amounts of irrelevant context, which can mislead the model and diminish its ability to focus on task-relevant information~\cite{hsieh2024ruler,kahatapitiya2024language,liu-etal-2024-lost,wu2025visualhaystack}.
However, the challenge extends beyond the \emph{size} of the context window to the more fundamental limitation that current models cannot \emph{utilize} it effectively~\cite{arnab2025temporalcot}. 
Most notably, with large context windows, models succumb to irrelevant distractors, highlighting a deeper attention failure rooted in the `bag-of-words' tendency in foundational spatial reasoning~\cite{qi2025beyoundsemantics}.
The problem is compounded by a positional encoding breakdown~\cite{ge2025v2pe,shi2025causality}, which arises due to na\"ive handling of continuous video frames and discrete language tokens in a similar manner. 
In contrast to previous work, we focus on highlighting how these fundamental issues are a direct cause of failure in complex temporal tasks like emotion understanding, along with an inference strategy with enriched context. 

A complementary line of work improves affective recognition by pairing facial-language models with structured or specialist facial cues. 
For DFER, DK-CLIP\cite{li2024dkclip} and PE-CLIP~\cite{saadi2025peclip} inject AU-grounded textual prompts into CLIP, Fine-CLIPER\cite{chen2024finecliper} fuses landmarks, segmentation masks, and MLLM-generated descriptions, and AU-DFER~\cite{liu2025audfer} adds AU–expression priors to conventional networks. 
For static FER, ExpLLM~\cite{lan2025expllm} derives a chain-of-thought from detected Action Units to guide an MLLM. Dedicated facial-video MLLMs such as
FaVChat\cite{zhao2025favchat} further report strong DFER results, surpassing a 72B generalist VLM on DFEW by a wide margin. Rather than contradicting our analysis, these works reinforce it: each succeeds by supplying generalist VLMs with the additional structure or specialisation our diagnosis shows they lack.

\subsection{Long-Tail Bias}
Long-tailed data distributions are known to degrade deep learning performance ~\cite{zhang2023deepsurvey-longtail,fang2023revisiting-longtail}, with bias embedding deeply in the feature representations, not just the final predictions~\cite{liu2020deep-longtail}.
Therefore, contemporary solutions aim to mitigate the final classification output and also seek to correct the underlying distortion in the feature space. 
Long-tail distribution bias is not limited to classification; it also prevails in regression~\cite{zhu2024regression-longtail}, semantic segmentation, and instance segmentation~\cite{wang2023balancing-longtail,zhang2021distribution-longtail}. 
A standard approach for handling long-tail imbalance is through class rebalancing~\cite{wongvorachan2023comparison-longtail,estabrooks2004multiple-longtail,liu2008exploratory-longtail}, which alters the training distribution by over-sampling tail classes~\cite{chawla2002smote} or under-sampling head classes. 
While simple, these methods risk overfitting on the tail or discarding valuable information from the head. 
Information augmentation techniques are also used to enrich the sparse tail class by creating synthetic samples using generative models like GANs~\cite{goodfellow2020gan,rangwani2021classgan,khorram2024taming-gan} or LLMs~\cite{wang2024llm-longtail,cloutier2023fine-llm-longtail}. 
Such approaches typically incur substantial computational overhead and are currently infeasible for video-based classification tasks using VLMs, which are already pre-trained on massive web-scale data. 
Alternative prior work decouples training into a two-stage approach~\cite{sun2025rethinking-longtail,kang2019decoupling-longtail,nam2023decoupled-longtail} in which a generalised feature extractor is trained firstly on a naturally imbalanced dataset, followed by classifier finetuning using a class-balanced sampler. 
We adopt pretraining plus class-balanced finetuning to probe the long-tail hypothesis with minimal confounds, deferring more advanced mitigation (\eg logit adjustment, reweighting, tail augmentation) to avoid added complexity and compute while keeping the focus on diagnosis.
\section{Systematic Diagnosis of VLM Deficiencies}
\label{sec:methodology}
We designed experiments to validate our hypothesis: \emph{class imbalance in training data and architectural limitations of VLMs lead to poor performance on temporal understanding tasks, like emotion classification}. We perform multi-class emotion classification using both VLMs and vision-only classifiers towards thorough experimental evaluation; experimental details follow. 

\subsection{Dataset and Evaluation Metrics}
\label{subsec:dataset}
To evaluate dynamic emotion recognition, we utilize two in-the-wild datasets: MAFW~\cite{liu2022mafw} (11-class emotions) and DFEW~\cite{jiang2020dfew} (7-class emotions). Standard DFER evaluation conventionally relies on two primary metrics: Weighted Average Recall (WAR), representing overall accuracy by accounting for class proportions, and Unweighted Average Recall (UAR), which averages recall across all classes regardless of frequency. UAR is uniquely critical as it treats every emotion category equally, preventing the majority classes from masking minority class failures.
Standard WAR metrics often mask failures on the minority tail in imbalanced datasets like DFEW. We therefore utilise balanced testing splits, sampling 495 videos (45 videos per class) for MAFW and 700 videos (100 videos per class) for DFEW, to ensure a rigorous evaluation free from majority-class bias and use the remaining videos for training. Due to this uniformity, UAR and WAR are mathematically identical in our evaluations.

\subsection{Models}\label{subsec:models}
We used representative examples from four model categories for our experimental work: closed-source general-purpose VLMs (Gemini2.5-Flash~\cite{comanici2025gemini}), open-source general-purpose VLMs (Qwen2.5-VL~\cite{Qwen2.5-VL}, Qwen2.5-Omni~\cite{xu2025qwen2omni}, Qwen3-VL~\cite{bai2025qwen3-vl}, Video-LLaVA~\cite{lin2024video-llava}, LLaVA-NeXT-Video~\cite{zhang2024llavanextvideo}, InternVL-3.0~\cite{zhu2025internvl3}), open-source task-specific VLMs (EmotionQwen~\cite{huang2025emotionqwen}), and open-source task-specific vision-only classifiers (MAE-DFER~\cite{sun2023maedfer} and HiCMAE~\cite{sun2024hicmae}). The task-specific models are models designed specifically for emotion understanding.
Where possible, open-source VLM parameter counts were kept within the 7--8B range for fair comparison.

\subsection{Long-tail distribution effects}
\label{subsec:long-tail-data-distribution}
\textbf{Real-world data distributions with respect to model performance:} Empirical category frequencies in many naturalistic and web-scale datasets are heavy-tailed and well approximated by Pareto and Zipf-like distributions~\cite{reed2001pareto,tang2020long}: a small number of object categories are exceptionally common, while the vast majority are rare. 

Early in the deep-learning era, computer vision models were often evaluated on artificially balanced datasets like ImageNet~\cite{deng2009imagenet} and MS-COCO~\cite{lin2014mscoco}, whereas modern VLMs are trained on larger, web-scraped data. Consequently, the underlying data distribution is often opaque, particularly the class-conditional sampling frequencies and long-tail coverage relative to any chosen label set. A class distribution is considered long‑tailed when a small minority of head classes account for a substantial share of all training examples, while the majority of tail classes each have few examples and their frequencies decline in a heavy‑tailed (often power‑law) manner. 
In practice, the analysis of common large-scale datasets, used for fine-tuning VLMs, exhibit long-tail patterns~\cite{parashar2024neglected}.
\vspace{-0.7cm}
\begin{figure}[htbp]
  \centering
  \includegraphics[width=\linewidth]{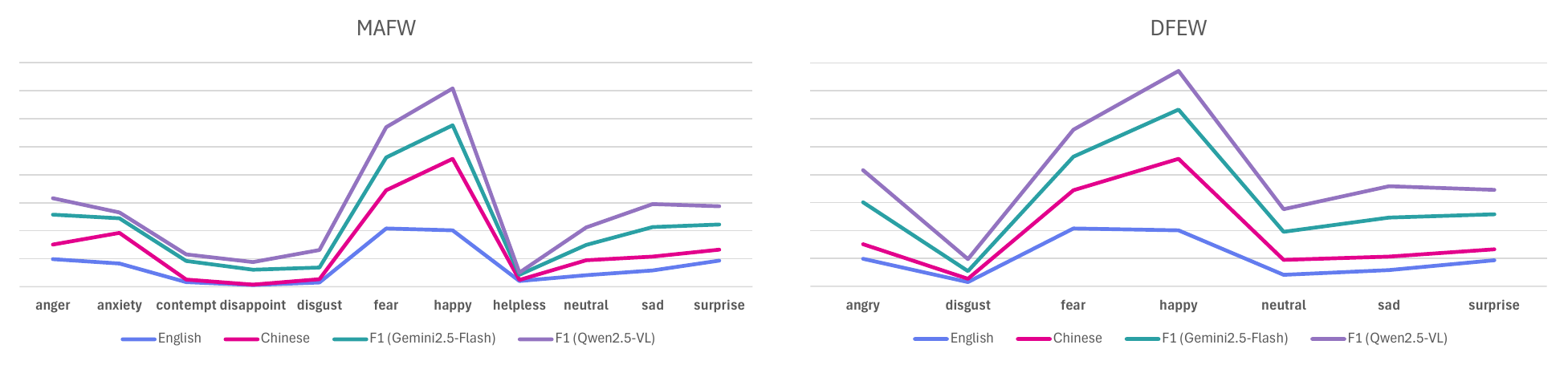}
  \vspace{-0.6cm}
  \caption{Correlation between Lexical Frequency and VLM Accuracy. We plot per-class F1 scores for VLMs against the lexical frequency of emotion terms in the Google Books Ngram corpus. The positive correlation suggests a head-class bias: rarer emotions (e.g., `contempt', `helplessness') have lower lexical frequency and correspond to weaker VLM performance.}
  \label{fig:ngram_vs_accuracy}
\vspace{-0.8cm}
\end{figure}

As an initial probing analysis, we estimate the data distribution for eleven classes from the MAFW~\cite{liu2022mafw} dataset. 
For VLMs trained on image-text pairs, we note that text frequency dictates supervision. With proprietary pre-training data unavailable, we utilize historical lexical frequencies in Google Books Ngrams~\cite{lin2012googlengram,michel2011googlengram} as a representative proxy for this `linguistic bottleneck'. As established in our results, the strong correlation between Ngram frequency and VLM error rates provides suggestive empirical support for this proxy. 
We selected the per-class frequency from the  July 2024 corpora for each emotion label in English and Chinese languages.
We compare Ngram frequency with the per-class F1 score of Gemini2.5-Flash~\cite{comanici2025gemini} and Qwen2.5-VL~\cite{Qwen2.5-VL} inference results, using a uniformly balanced test set from MAFW~\cite{liu2022mafw}, in a zero-shot setting (Figure~\ref{fig:ngram_vs_accuracy}). 
We calculate the Pearson correlation coefficient for each pair to estimate the relationship between model predictive accuracy and the occurrence of each class in Google Books Ngram data. 
Unlike prior work establishing long-tail inheritance for general objects via synonym counts in pretraining captions~\cite{parashar2024neglected}, we show this bias in emotion recognition couples to historical lexical prevalence, with cross-lingual consistency suggesting a deeper cultural and linguistic skew beyond raw pretraining statistics.
The Pearson correlation coefficient is calculated using:
\begin{equation}
r = \frac{\sum_{i=1}^{n} (x_i - \bar{x})(y_i - \bar{y})}{\sqrt{\sum_{i=1}^{n} (x_i - \bar{x})^2 \times \sum_{i=1}^{n} (y_i - \bar{y})^2}},
\label{eq:pearson-correlation}
\end{equation}
\begin{table}[htbp]
\centering
\setlength{\tabcolsep}{6pt}     
\begin{tabular}{l c c c c}
\hline
 & \thead{\textbf{English}} 
 & \thead{\textbf{Chinese}} 
 & \thead{\textbf{Gemini2.5-Flash}} 
 & \thead{\textbf{Qwen2.5-VL}} \\
\hline
\textbf{English} 
 & \makecell{1.000\\(---)} 
 & \makecell{0.8825\\(0.0003)} 
 & \makecell{0.7927\\(0.0036)} 
 & \makecell{0.8041\\(0.0029)} \\
\hline
\textbf{Chinese} 
 & \makecell{0.8825\\(0.0003)} 
 & \makecell{1.000\\(---)} 
 & \makecell{0.6343\\(0.0361)} 
 & \makecell{0.7547\\(0.0073)} \\
\hline
\textbf{Gemini2.5-Flash}
 & \makecell{0.7927\\(0.0036)} 
 & \makecell{0.6343\\(0.0361)} 
 & \makecell{1.000\\(---)} 
 & \makecell{0.8176\\(0.0021)} \\
\hline
\textbf{Qwen2.5-VL}
 & \makecell{0.8041\\(0.0029)} 
 & \makecell{0.7547\\(0.0073)} 
 & \makecell{0.8176\\(0.0021)} 
 & \makecell{1.000\\(---)} \\
\hline
\vspace{-0.2cm}
\end{tabular}
\caption{Correlation between per-class occurrence in Google Books Ngram data (English, Chinese) and F1 accuracy of Gemini2.5-Flash and Qwen2.5-VL on MAFW. We report correlation coefficients and ($p$-values).}
\label{tab:ngram-f1-pcorrelation}
\vspace{-1.5cm}
\end{table}
\\
where $x$ and $y$ are the two signals and $\bar{x}$ and $\bar{y}$ are their respective means, and $n$ is the number of data points (here emotion classes). 
Statistical significance is assessed using a two-tailed $t$-test, with $p$-values across model-language pairs (Gemini2.5-Flash and Qwen2.5-VL; English and Chinese) indicating that all correlations are statistically significant ($p < 0.05$; see Table~\ref{tab:ngram-f1-pcorrelation}). 
VLMs are trained predominantly on English web data, so English Ngrams are expected to be the closer proxy; Chinese affective lemmas are also more polysemous, inflating raw frequency relative to affective usage.The cross-lingual consistency of the correlation, rather than the absolute frequencies, is what strengthens the long-tail interpretation.

We further examine whether similar correlations emerge in specialist (vision-only) emotion classifiers, thus controlling for VLM‑specific pretraining and context windows. Essentially our objective is to test whether the observed accuracy–frequency coupling reflects the long‑tailed training distribution itself. 
We selected two state-of-the-art models MAE-DFER~\cite{sun2023maedfer} and HiCMAE~\cite{sun2024hicmae}, which are based on the MAE~\cite{he2022mae} architecture and pre-trained on Voxceleb2 data~\cite{Chung18voxceleb2}. Voxceleb2 contains no emotion classification labels yet we treat this as a large audio-visual resource, capable of producing a strong, general purpose feature extractor. 
We fine-tuned the models on the training split of the MAFW dataset (Sec.~\ref{subsec:dataset}), which exhibits class imbalance. 
Some of the classes are high-frequency: anger, happiness, neutral, sadness, surprise, while others are nuanced or long-tail: contempt, disappointment, helplessness. 
We evaluate the performance of the fine-tuned models on the evenly sampled testing split. 
We observe a similar performance pattern in specialized vision-only video classification models. For instance, the F1 score for happiness was 0.70 in MAE-DFER, while low-frequency classes like contempt and disappointment has 0.18 and 0.22, respectively.
Similarly, HiCMAE has an F1 score of 0.69 for happiness, but 0.04 and 0.16 for helplessness and disappointment, respectively. 
These findings are consistent with our conjecture that web-scale pretraining data, ingested by VLMs, are long‑tailed over emotion categories, contributing to weaker performance on under‑represented emotions. 

\textbf{Validating the performance on a class-balanced dataset:} 
As Gemini2.5-Flash cannot be fine-tuned (code unavailable), we report its per-class behaviour as a measurement of bias rather than a mitigation result. This is relevant for downstream users of closed models, who cannot retrain but can use such estimates to calibrate trust.
The model achieved F1 scores of 0.60 for happiness, 0.09 for helplessness, and 0.26 for disappointment. These results are comparable to those obtained with Qwen2.5-VL, which recorded F1 scores of 0.54 for happiness, 0.0 for helplessness, and 0.07 for disappointment. This indicates that current VLMs have not effectively learned to recognize nuanced emotional expressions; instead, they tend to rely on memorization of the most frequent emotion labels.
\begin{figure*}[t]
  \centering
  \includegraphics[width=\linewidth]{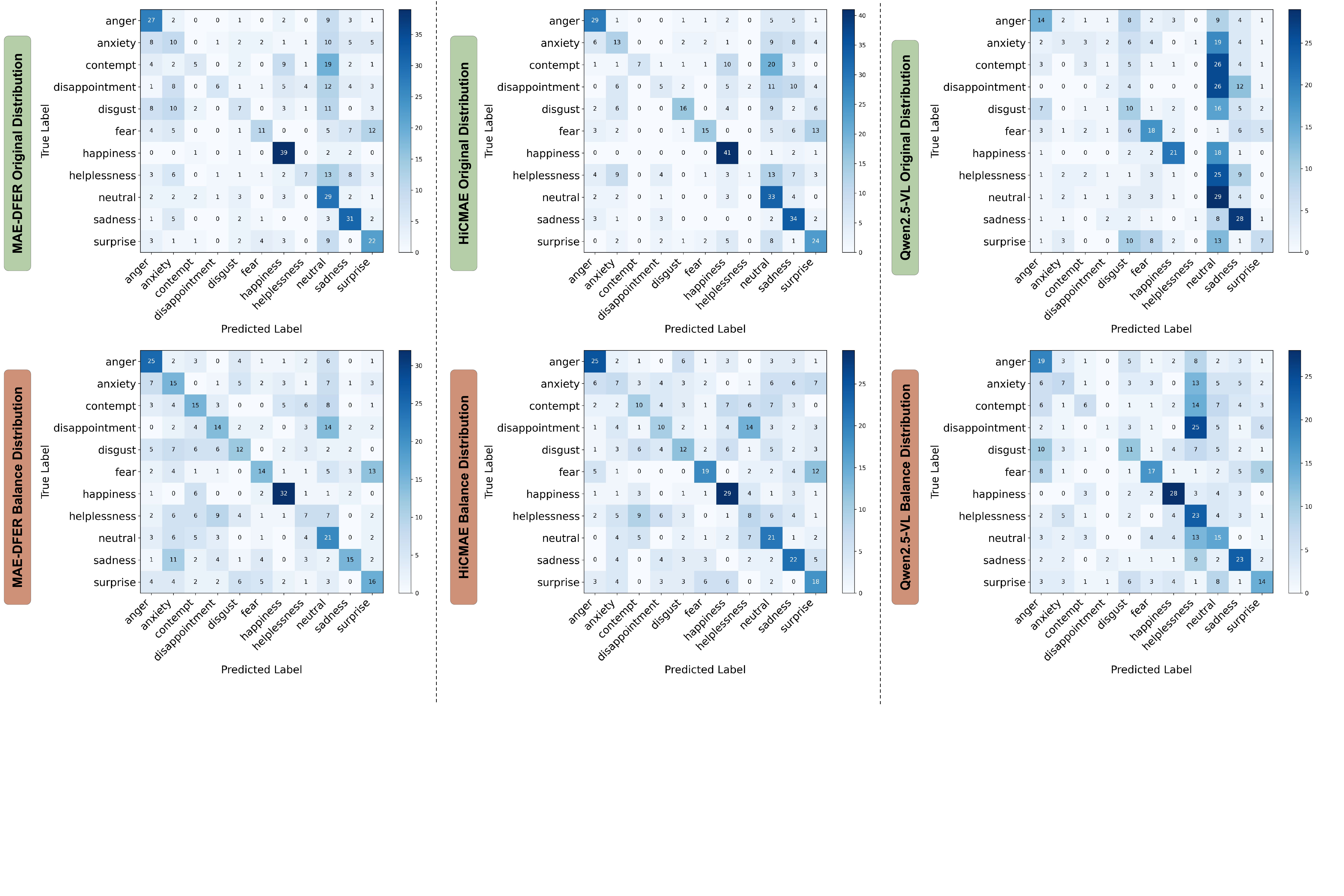}
  \caption{Confusion matrices for the classification task on MAFW dataset. Original data distribution (top row); fine-tuning with balanced dataset distribution (bottom row).}
  \label{fig:confusion_matrix_balance_sampling}
  \vspace{-0.7cm}
\end{figure*}

When presented with videos depicting rare emotions such as disappointment, the investigated models often resort to predicting the high-frequency or common emotion categories (Figure.~\ref{fig:confusion_matrix_balance_sampling}, \eg top-row, last-column). The confusion matrices reveal strong diagonals for high-frequency classes (\eg, neutral), while predictions for long-tail emotions (\eg, contempt, disappointment) are scattered and inconsistent. 
This pattern refutes the assumption that class confusion may stem solely from semantic similarity (e.g., Sadness vs.~Disappointment). 
Crucially, our error analysis reveals \emph{asymmetric} confusion: rare classes systematically collapse into common classes (Disappointment $\rightarrow$ Sadness), but the reverse almost never holds. Pure semantic confusion is bidirectional, whereas this pronounced asymmetry confirms that models default to high-frequency priors rather than failing due to semantic ambiguity. We formalise this with a Directional Asymmetry Index in the supplementary material.
Overall, the results highlight a key limitation of current VLMs: training on long-tailed datasets leads to the systematic collapse of low-frequency emotions toward high‑frequency categories, thereby diminishing sensitivity to subtle emotional variations.

\subsection{Temporal Understanding}\label{subsec:temporal-understanding}
\textbf{Do VLMs leverage frame ordering information?} 
Emotion recognition relies on temporal cues that involve micro- and macro-expressions~\cite{haggard1966micromomentary,ekman1969nonverbal-mircoexpression,matsumoto2011evidence-microexpression}, evolving over time. 
This suggests that temporal information likely plays a key role.
To examine how crucial temporal information is to task success, we modify the input video frame ordering. 
For a given input video $V$ with $[I_1,I_2,\ldots,I_L]$ frames, we randomly shuffle the frame order. 
We evaluate the performance of both VLMs and vision-only classifier models with two sets of input videos containing identical visual (\ie spatial and appearance) information: one with the original chronological temporal ordering and one with the temporal frames shuffled (FS). 
The specialized temporal models (MAE-DFER, HiCMAE) show a significant (\textbf{15--16\%}) performance drop in macro-averaged F1 score.
In contrast, VLM model performance remains largely indifferent to frame shuffling (see Table~\ref{tab:frame-shuffling}). 
Across varying model scales and specific visual integration techniques, the vulnerability remains strikingly consistent: VLMs performance remains largely indifferent to random temporal shuffling, confirming an order-agnostic `bag-of-frames' processing strategy in all instances.

Our experiment highlights a critical limitation in the temporal awareness and reasoning of current VLMs. 
This insensitivity provides strong evidence of an over-reliance on frame-level spatial features, indicating a failure to reason across temporal information. We demonstrate that models are not merely failing to reason, but are architecturally defaulting to a `bag-of-frames' processing strategy.
Furthermore, this suggests that the order-aware processing imposed by causal attention masking, which is intended to implicitly guide inter-visual token interactions and guarantee architectural temporal ordering, is also failing. 
Our frame-shuffling probe provides a real-world corollary that complements SpookyBench~\cite{upadhyay2026timeblindness} by evaluating naturalistic, high-resolution video rather than noise-based patterns, holding spatial content constant while removing only temporal order. While both works find an over-reliance on spatial cues, we identify a distinct failure mode: redundancy-driven attentional dilution. Unlike the failure on noise sequences, we show that accuracy drops as visual token density increases and saturates the model budget, revealing an architectural inability to isolate fleeting signals from clear but redundant frames.
\begin{table}[htbp]
\centering
\setlength{\tabcolsep}{5pt}
\begin{tabular}{l  c c c | c c c }
\hline
\multirow{2}{*}{\textbf{Model}} & \multicolumn{3}{c|}{\textbf{MAFW}} & \multicolumn{3}{c}{\textbf{DFEW}} \\
\cline{2-7}
 & \textbf{Precision} & \textbf{Recall} & \textbf{F1} & \textbf{Precision} & \textbf{Recall} & \textbf{F1} \\
\hline
MAE-DFER & 0.4394 & 0.3919 & 0.3602 & 0.6883 & 0.6086 & 0.5645  \\
MAE-DFER FS & 0.4959 & 0.3354 & 0.3041 & 0.5006	& 0.5229	& 0.4802  \\
\hline
HiCMAE & 0.5040 & 0.4404 & 0.3993 & 0.6935 & 0.6214 & 0.5725  \\
HiCMAE FS & 0.4979 & 0.3778 & 0.3345 & 0.4807 & 0.5329 & 0.4797  \\
\Xhline{2pt}
Qwen2.5-VL & 0.2849 & 0.2727 & 0.2449 & 0.5053 &  0.4614 & 0.4552 \\
Qwen2.5-VL FS & 0.3527 & 0.2727 & 0.2506 & 0.5015 & 0.4643  & 0.4534 \\
\hline
Qwen2.5-Omni & 0.4517 & 0.3253 & 0.306 & 0.5557 & 0.4700	& 0.4296 \\
Qwen2.5-Omni FS & 0.4502 & 0.3172 & 0.2972 & 0.5276 & 0.4614	& 0.4226 \\
\hline
Qwen3-VL & 0.4180 & 0.3313 & 0.2738 & 0.6380 & 0.5843 & 0.5511 \\
Qwen3-VL FS & 0.3198 & 0.3273 & 0.2615 & 0.6468	& 0.5871	& 0.5538 \\
\hline
EmotionQwen & 0.3478 & 0.2925 & 0.2581 &  0.5521 & 0.5329 & 0.5010  \\ 
EmotionQwen FS & 0.3185 & 0.3057 & 0.2517 & 0.5383 & 0.4968 & 0.4895  \\ 
\hline
Video-LLaVA & 0.1126 & 0.1630 & 0.0870 & 0.1490 & 0.2800 & 0.1654\\
Video-LLaVA FS & 0.0811 & 0.1569 & 0.0814 & 0.1514 & 0.2570 & 0.1531 \\
\hline
LLaVA-NeXT-V & 0.1853 & 0.2040 & 0.1438 & 0.5351 & 0.3474 & 0.2969 \\
LLaVA-NeXT-V FS & 0.1705 & 0.1980 & 0.1382 & 0.4366 & 0.3314 & 0.2712 \\
\hline
InternVL-3.0 & 0.3441 & 0.2828 & 0.2445 & 0.5561 &  0.5357 & 0.5044 \\
InternVL-3.0 FS & 0.3370 & 0.2667 & 0.2326 & 0.5669 & 0.5214 & 0.4985 \\
\hline
Gemini2.5-Flash & 0.4112 & 0.3869 & 0.3758 &  0.6412 & 0.6331 & 0.6008  \\
Gemini2.5-Flash FS & 0.4286 & 0.3817 & 0.3626 &  0.6246 & 0.6120 & 0.5778  \\
\hline
\vspace{0.01cm}
\end{tabular}
\caption{Model performance on two datasets (MAFW and DFEW) for normal videos vs Frame Shuffled (FS) videos using vision-only classifiers (top section) and VLM-based architectures (bottom section).}
\label{tab:frame-shuffling}
\vspace{-1.0cm}
\end{table}
\\\\
\textbf{Context Windows and Irrelevant Distractors:} Contemporary VLMs process visual frames by dividing them into tokens. These visual tokens are treated analogously to text tokens within the model architecture. 
Although this design choice appears straightforward and easily adaptable, it overlooks the fundamental distinctions between the modalities.
Text tokens are discrete semantic units and categorical: equality is all‑or‑nothing. Thus `cat' and `dog' are distinct labels; in a standard one‑hot representation, their vectors are orthogonal. 
Video frames, however, are dense, continuous, and highly correlated; two adjacent frames in a video are often highly similar. This signifies that a large portion of video information is typically redundant and, for a specific targeted query, often even irrelevant. This problem is compounded by a mechanistic imbalance: video token embeddings possess a significantly larger vector norm than text token embeddings~\cite{qi2025beyoundsemantics}.  
Given that attention mechanisms already struggle to model extended temporal context, enlarging the context with distracting and irrelevant content can further exacerbate this limitation. 
This phenomenon is a video-centric manifestation of the `lost-in-the-middle' problem~\cite{liu-etal-2024-lost}, where model attention is confused by noise, and performance degrades as sequence length increases. 

\begin{figure}[htbp]
\centering
\includegraphics[width=0.5\linewidth]{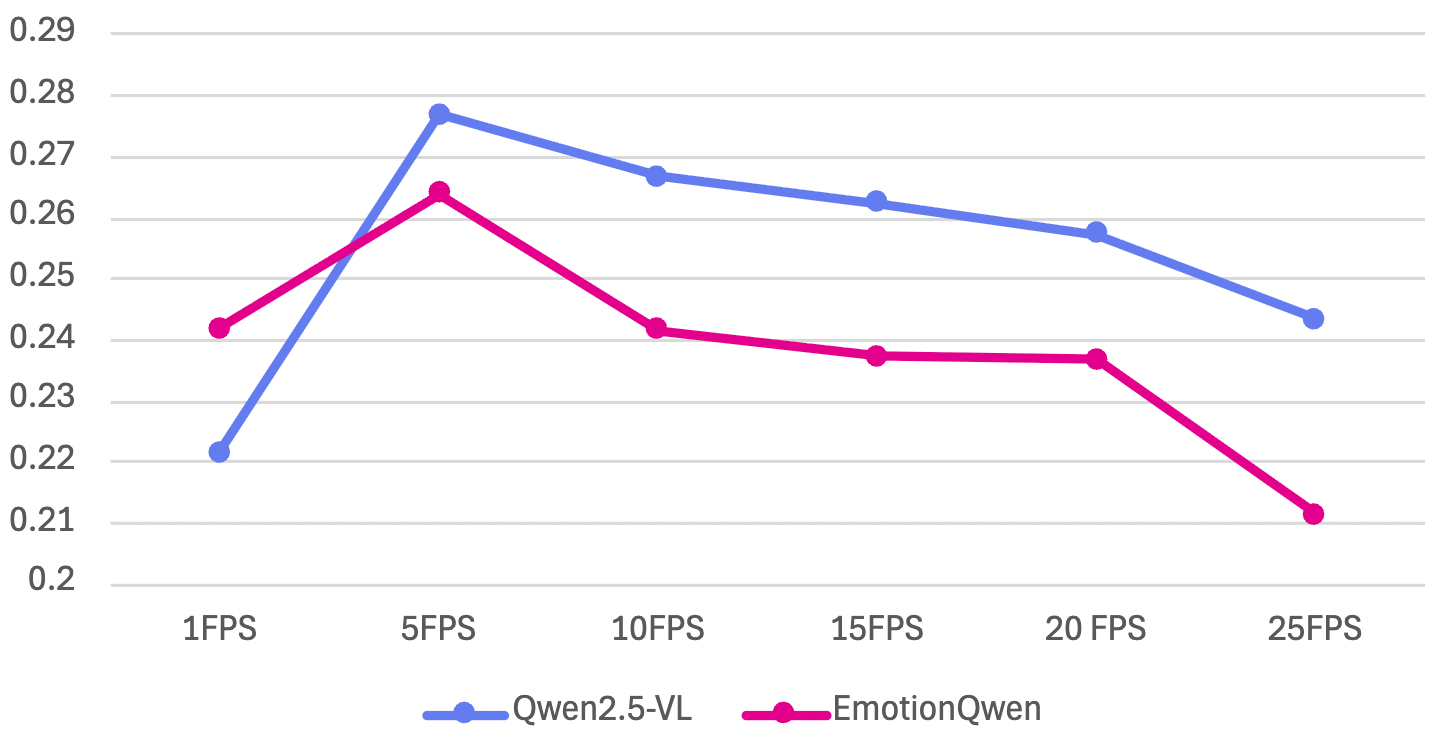}
\vspace{-0.2cm}
  \caption{VLM Performance vs. Input Frame Rate. We plot the macro-averaged F1-score of Qwen2.5-VL and EmotionQwen as a function of frames-per-second (FPS) on MAFW dataset. The quasi-bell-shaped curve shows performance initially improving (1--5 FPS) but then degrading significantly 
  ($>5$ FPS), demonstrating attentional dilution from redundant visual tokens.}
  \label{fig:fps-variation}
  \vspace{-0.6cm}
\end{figure}

To evaluate the performance of the model with a varying amount of visual information, we perform an experiment that controls the number of video input frames. We perform this experiment on Qwen2.5-VL~\cite{Qwen2.5-VL} and EmotionQwen~\cite{huang2025emotionqwen}, as their open-source nature allows for fine-grained control over frame-sampling inputs. Closed-source models like Gemini2.5-Flash~\cite{comanici2025gemini} do not provide this level of API control, and are thus omitted from this specific test.
We started by sampling one frame every second from video input and instruct the Qwen2.5-VL~\cite{Qwen2.5-VL} model to infer the emotional state from the given sampled frames only. We then increase the number of frames per second to 5, 10, 15, 20, and 25. We observe that, initially, model performance improves when it enjoys access to more frames \ie, more visual information is available to infer the emotional state. However, as we increase the number of frames, model performance starts to degrade, and even eventually drops below the performance obtained when using one frame per second. 
We repeat the experiment with  EmotionQwen~\cite{huang2025emotionqwen} and note similar performance trends. (Additional results in supplementary material)

The performance curves shown in Figure~\ref{fig:fps-variation} provide empirical evidence in support of our hypothesis that attention mechanisms in VLMs degrade with excessive information and fail to retain long-term context. The initial performance gain is expected, and the subsequent degradation at high frame rates is a notable finding that highlights underlying architectural limitations. 
This behavior is particularly damaging for complex temporal tasks like emotion understanding, which likely require fine-grained temporal sensitivity to reliably classify micro-expressions. 
The quasi-bell-shaped curve (Figure~\ref{fig:fps-variation}) provides compelling empirical evidence that VLM attention is not just limited, but is actively diluted by high-density, redundant visual input. This confirms that redundancy and distractors, and a limited ability to exploit long context, constitute a principal bottleneck for temporal reasoning in contemporary VLMs.

\section{Enhancing Emotions via Distribution and Dynamics}
\label{sec:experiments}
\subsection{Mitigating Long-Tail Bias} \label{subsec:mitigate-long-tail}
Furthering our high-level correlative analysis (Section~\ref{subsec:long-tail-data-distribution}), we next consider direct empirical validation of the long-tail hypothesis. To evidence that data imbalance is a primary cause of failure, we employed a `decoupled' training strategy.
We take advantage of the strong feature extraction capabilities learned by models trained on large-scale data. We sampled ${\sim}$1500 videos from the MAFW dataset, such that each class has an equal number of samples. We fine-tune both MAE-DFER and HiCMAE on the new uniformly sampled balanced distribution. 
Despite a six-fold reduction in available fine-tuning data, the distribution of predicted labels at inference is appreciably more uniform across classes, indicating reduced head-class bias. 
We repeat a similar experiment with Qwen2.5-VL, using LoRA-based finetuning~\cite{hu2022lora}. 
As shown in Figure~\ref{fig:confusion_matrix_balance_sampling}, this balanced finetuning validates our long-tail hypothesis. It improves tail-class performance and yields a more uniform prediction distribution, confirming the issue is data bias rather than an inherent incapacity for these emotions.
In summary, our decoupled pretrain--then--balanced finetuning is simple yet effective: it preserves web-scale representations while correcting head-class bias without introducing additional modeling or data-generation confounds. This lightweight step yields consistent gains across all models, providing a pragmatic and clean mitigation when the pretraining distribution is opaque and retraining is infeasible.

\begin{figure*}[tp]
\centering\includegraphics[width=\linewidth]{images/msce_architecture.pdf}
  \caption{Multi-Stage Context Enrichment (MSCE) is a two-stage inference pipeline. Top (baseline): a VLM given only sparse keyframes misses the fleeting micro-expression in the in-between frames and predicts \emph{Neutral}. Bottom (MSCE): in Stage 1 (Motion-to-Text Translation), the VLM converts the in-between frames into a compact motion description of the micro-expression; in Stage 2 (Interleaved Context Classification), this textual summary is interleaved with the sparse keyframes for a final prediction, recovering the correct label (\emph{Happiness}).}
  \label{fig:architecture}
  \vspace{-0.4cm}
\end{figure*}

\subsection{Multi-Stage Context Enrichment for Temporal Reasoning}
Our analysis in Section~\ref{subsec:temporal-understanding} provides empirical evidence that VLMs struggle with temporal reasoning due to two competing problems: 
1) sparse sampling (\eg, 1 FPS) misses fleeting micro-expressions, and 2) dense sampling (\eg $>$10  FPS) overwhelms context windows with redundant distractions. 
This results in a critical problem: diagnostically salient affective cues frequently reside in frames that sparse sampling necessarily excludes under context‑window constraints. 
Unlike frame selection methods such as TCoT~\cite{arnab2025temporalcot}, which discard content in unsampled temporal gaps and risk losing information, we propose a Multi-Stage Context Enrichment (MSCE) strategy. Rather than discarding these gaps, MSCE translates their inherent motion cues into compact natural-language summaries, a modality that vision-language models (VLMs) process highly effectively. 
By converting high-volume, redundant visual tokens into low-volume, information-dense text tokens, we augment the model’s effective context while respecting the token budget. 
This strategy is illustrated in Figure~\ref{fig:architecture} and is executed in two stages.
\\\\
\textbf{MSCE Stage 1: Motion-to-Text Translation}: For video $V$, we sample $n$ frames $K = \{k_1, k_2,\ldots,k_n\}$, corresponding to our baseline sparse sampling strategy (e.g., 1 FPS dictates that $k_1$ is at 0.0 seconds, $k_2$ is at 1.0 seconds, etc.). We refer to this as a set of keyframes and this sparse sampling creates $n-1$ temporal gaps, containing unobserved intermediate frames. For each temporal gap $j$ (containing frames between $k_i$ and $k_{i+1}$), we sample a small, denser set of intermediate frames, $G_j = \{g_{j_1}, g_{j_2},\ldots,g_{j_m}\}$. These frames are sampled uniformly, to maximize input information. The set $G_j$ is intentionally kept small ($m=4$ in our experiments) to avoid overwhelming VLMs and minimize attention dilution.  
The set $G_j$ are then processed through the VLM to provide a textual summary of the facial temporal movements, and to capture micro-expressions. This granular text, containing motion information, serves as a surrogate for the temporal information lost under sparse frame sampling, and provides high-level context, absent from the visual inputs. 
At the end of Stage 1, we have a list of $n-1$ textual summaries, $T = \{t_1, t_2,..., t_{n-1}\}$, where each $t_i$ describes motion that the sparse keyframes $K$ have missed.
\\\\
\textbf{MSCE Stage 2: Interleaved Context Classification}: The second stage synthesizes sparsely sampled frames $K$ with the generated textual motion summaries to provide a holistic emotion classification. We create a single text prompt that inter-leaves the keyframes from $K$ with the text summaries from $T$ in chronological order. The input to the final VLM is structured as:\\
\mbox{$\{k_1, t_1, k_2, t_2, \ldots, k_{n-1},$ $t_{n-1}, k_n\}$}. The VLM is then prompted to make a final classification based on this complete, enriched context using prompt:

\begin{tcolorbox}[enhanced, colback=cyan!10, colframe=cyan!50!black, boxrule=0.5pt, arc=1mm, left=1mm,right=1mm,top=1mm,bottom=1mm, opacityback=0.5, opacityframe=0.1]
\noindent \texttt{Analyze the following sequence of sparse keyframes and the detailed motion descriptions for the gaps between them. 
\\FrameID 1:\{$k_1$\}, Motion:\{$t_1$\},$\cdots$, FrameID $n$:\{$k_n$\}. 
\\Question: \{question\}
\\Emotion Labels choices: \{List of Emotion Labels\}
}
\end{tcolorbox}
\vspace{-0.5cm}
\begin{table}[htbp]
\centering
\setlength{\tabcolsep}{3pt}
\begin{tabular}{l ccc | ccc}
\hline
 & \multicolumn{3}{c|}{\textbf{MAFW}} & \multicolumn{3}{c}{\textbf{DFEW}} \\
\cline{2-4} \cline{5-7}
\textbf{Model} & \textbf{Precision} & \textbf{Recall} & \textbf{F1} & \textbf{Precision} & \textbf{Recall} & \textbf{F1} \\
\hline
Qwen2.5-VL & 0.2849 & 0.2727 & 0.2449 &  0.5053 & 0.4614 & 0.4552  \\
Qwen2.5-VL +MSCE & 0.3293 & 0.3091 & 0.2731 &  0.5204 & 0.4900 & 0.4820  \\
\hline
EmotionQwen & 0.3478 & 0.2925 & 0.2581 & 0.5521 & 0.5329 & 0.5010  \\ 
EmotionQwen +MSCE & 0.3617 & 0.3061 & 0.2683 &  0.5635 & 0.5429 & 0.5147  \\ 
\hline
LLaVA-NeXT-V & 0.1853 & 0.2040 & 0.1438 & 0.5351 & 0.3474 & 0.2969  \\
LLaVA-NeXT-V +MSCE & 0.2294 & 0.2323 & 0.1715 &  0.5167 & 0.3714 & 0.3171  \\
\hline
\vspace{0.001cm}
\end{tabular}
\caption{Our Multi-Stage Context Enrichment (MSCE) strategy results in consistent F1 score improvement compared with baseline sparse sampling.}
\label{tab:context-enrichment}
\vspace{-0.9cm}
\end{table}

This interleaved architecture directly remedies the core failures identified in Section~\ref{subsec:temporal-understanding}. 
We further contrast MSCE with the frame-selection method TCoT~\cite{arnab2025temporalcot} under a matched Qwen2.5-VL setup. Because micro-expressions last only
$0.25$--$0.5$\,s, a coarse macro-level selection search such as TCoT~\cite{arnab2025temporalcot} discards the unselected gaps that carry vital affective signal, leaving its F1 at or below the sparse baseline ($0.241$ vs.\ $0.245$ on MAFW, $0.450$ vs.\ $0.455$ on DFEW), whereas MSCE improves over both ($0.273$ and $0.482$).
Our strategy yields a consistent performance improvement (Table~\ref{tab:context-enrichment}), further demonstrating that reasoning over temporally intermediate content is critical.
While these gains are moderate, they are diagnostically conclusive for our temporal hypothesis. 
They empirically prove that critical affective cues reside in the temporal gaps discarded by sparse sampling, and demonstrate that these cues can be recovered and utilized by VLMs via modality translation. 
The textual summary $t_i$ acts as a `temporal bridge', providing explicit, high-level semantic context that spans the visual gap between keyframe $k_i$ and $k_{i+1}$. This allows the model to reason over the evolution of an expression, \eg a micro-expression fleetingly appearing and disappearing, and consequently, it moves beyond the order-agnostic aggregation of per-frame spatial features. 
We provide initial evidence that modality conversion can preserve essential temporal information while avoiding visual‑token overload.
\vspace{-0.1cm}
\section{Limitations and Discussion}
\label{sec:limitation}
While MSCE provides measurable gains, its implementation highlights three structural frontiers in long-form video understanding. 
First, MSCE functions as a strategic intervention rather than a final resolution. 
While it improves access to temporal cues, we acknowledge that text summarization may introduce generation noise. However, this dependency strategically shifts the burden from the model's weak long-term temporal retention to its strong short-term descriptive capabilities. By interleaving text summaries with original keyframes, the text serves as a supporting signal rather than a sole dependency, with empirical results confirming that the recovered temporal information outweighs any noise introduced by the process. Future work may pursue video-native architectures to permanently resolve the underlying deficit in long-horizon temporal modeling.
Second, current VLM attention degrades with very long visual contexts, which constrains sensitivity to micro-expressions; potential remedies include hierarchical or sparse temporal attention and improved long-sequence positional encodings. 
Third, our long-tail analysis relies on Google Books Ngram frequencies as a correlative, non-causal, proxy for training priors. Given the opacity of web-scale corpora, additional frequency proxies can strengthen the analysis.
\section{Conclusion}
\label{sec:conclusion}
We investigate the critical failure of modern VLMs to comprehend human emotion and empirically demonstrate that this problem stems from two fundamental flaws. 
First, we provide quantitative evidence that VLM failures are data-driven. By correlating model accuracy with lexical frequency, we show that VLMs inherit a long-tail bias from their pretraining data, causing them to conflate rare emotions with their high-frequency counterparts. 
Second, we demonstrate that this limitation is inherently architectural. Empirically, VLMs struggle with temporal modeling, lacking sensitivity to frame order. Moreover, performance deteriorates as visual input density increases, indicating limited capacity to exploit the dense context required to detect micro-expressions. 
Our experiments suggest that these shortcomings, while potentially addressable, are deeply rooted in current data and architectural practices. 
Consequently, advancing toward robust affective computing will require fundamental shifts, including a thorough reconsideration of VLM training regimes and temporal modeling architectures.

%
%
\bibliographystyle{splncs04}
\bibliography{main}

\clearpage
\begin{center}
{\Large\bfseries Why Do Vision Language Models Struggle To Recognize Human Emotions? \par}
\vspace{0.9em}
{\large Supplementary Material}
\end{center}
\appendix

\captionsetup[table]{skip=6pt}

\setcounter{table}{0}
\renewcommand{\thetable}{S\arabic{table}}
\setcounter{figure}{0}
\renewcommand{\thefigure}{S\arabic{figure}}
\setcounter{equation}{0}
\renewcommand{\theequation}{S\arabic{equation}}
\setcounter{section}{0}
\renewcommand{\thesection}{S\arabic{section}}

\section{Visual Prompting: modality study}
\label{supsec:ablation}

\begin{figure}[htbp]
\centering
\includegraphics[width=0.8\linewidth]{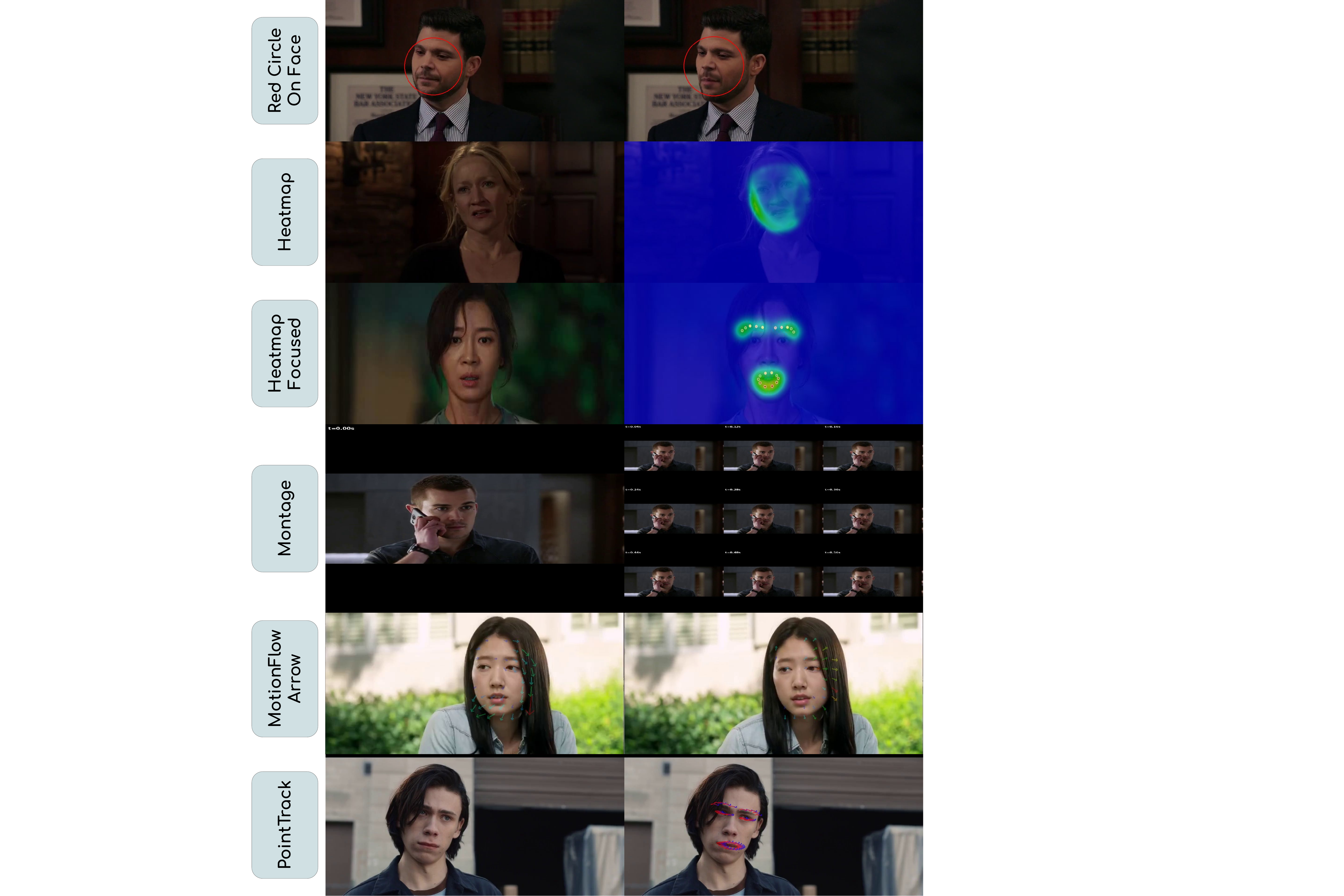}
\caption{Visualization of failed input modalities for emotion recognition. We test whether explicitly guiding VLM attention via visual prompting can bridge the temporal gap. Techniques such as Red Circle and Heatmap introduce visual artifacts that shift the input out of the model's training distribution, degrading performance. Video Montage reduces the resolution of individual frames, obscuring critical micro-expressions.}
\vspace{-0.7cm}
\label{fig:supp_ablation}
\end{figure}

We further explore performance improvement achievable by examining whether temporal limitations can be mitigated through explicit attention guidance, via visual feature engineering, rather than by enriching textual context. We conduct an empirical exploration of diverse visual prompting and feature-engineering strategies, varying the format of visual information and representations of the input to elicit improved performance. 
We experimented with six alternative input modalities (Figure~\ref{fig:supp_ablation}), widely used in computer vision:
\begin{itemize}
    \item 2D spatial annotations: imputing red circles to encourage model attention focus to facial expressions.
    \item Heatmap: overlay heatmaps to highlight the motion on the face region.
    \item Focused Heatmap: overlay heatmaps to highlight motion; specific regions (eyes and lips).
    \item Video Montage: concatenation of frames into a single grid-image to `spatialize' temporal information.
    \item MotionFlow Arrows: Overlaying optical flow vectors to explicitly encode motion directions.
    \item Point Tracking: Visualizing keypoint trajectories of lips and eyebrows.
\end{itemize}

\begin{table}[htbp]
\centering
\begin{tabular}{l c c c}
\hline
\textbf{Model} & \textbf{Precision} & \textbf{Recall} & \textbf{F1} \\
\hline
Baseline & 0.2849 & 0.2727 & 0.2449 \\
\hline
+ Red Circle On Face & 0.3168 & 0.2772 & 0.2385 \\
+ Heatmap & 0.3214 & 0.2788 & 0.2324  \\
+ Focused Heatmap & 0.3733 & 0.2727 & 0.2297 \\
+ Video Montage & 0.215 & 0.2444 & 0.2012  \\
+ MotionFlow Arrow& 0.3563 & 0.2103 & 0.1792 \\
+ PointTrack& 0.4413 & 0.2687 & 0.2401 \\
\hline
+ MSCE & 0.3293 & 0.3091 & 0.2731 \\
\hline
\end{tabular}
\caption{Sensitivity study varying the input modalities to the Qwen2.5-VL model. We compare the proposed MSCE strategy against various visual prompting techniques. All visual modifications degrade performance compared to the baseline sparse sampling, indicating that visual modifications introduce distribution shifts. MSCE (textual enrichment) yields F1 score improvement over the baseline. 
}
\label{tab:supp_ablation}
\end{table}

As shown in Table~\ref{tab:supp_ablation}, all visual modifications resulted in performance degradation compared to the baseline sparse sampling (F1: 0.2449). We attribute this to distribution shift. VLMs are pre-trained on natural images; artifacts like red circles, dense flow maps, or montages introduce visual noise that moves the input out-of-distribution (OOD) for the frozen visual encoder. Similarly, heatmaps introduce a new visual modality that proves hard to interpret by a pre-trained visual encoder. Leveraging the text modality in the MSCE approach avoids OOD visual artifacts while preserving the temporal motion information of `in-between' frames.

\section{Extended Analysis of Temporal Sensitivity (Frame Shuffling)}
\label{supsec:frame_shuffling}

We selected models to cover distinct architectural paradigms: (i) Proprietary Generalist VLMs (Gemini2.5-Flash) to establish a closed-source benchmark. (ii) Open-Source Generalist VLMs (Qwen2.5-VL) to establish a standard spatial reasoning baseline. (iii) Omni-modal Models (Qwen2.5-Omni) to test if native audio-visual integration improves temporal grounding. (iv) Task-Specific VLMs (EmotionQwen) designed explicitly for emotion understanding. (v) Video-Native VLMs (Video-LLaVA, LLaVA-NeXT-Video) to test models explicitly instruction-tuned on video data. (vi) Next-Gen Generalist VLMs (Qwen3-VL, InternVL-3.0) to test if recent architectural updates resolve the issue. Where possible, we kept the open-source VLM parameter sizes consistent to the 7-8B range for fair comparison. All experiments were performed on a single A100 (80GB) GPU.

\begin{table*}[htbp]
\centering
\small
\resizebox{\textwidth}{!}{\begin{tabular}{l | c c c c c c c c c c c | c}
\hline
\textbf{Model} &
\textbf{AG} &
\textbf{AX} &
\textbf{CO} &
\textbf{DA} &
\textbf{DG} &
\textbf{FE} &
\textbf{HA} &
\textbf{HP} &
\textbf{NU} &
\textbf{SD} &
\textbf{SU} &
\textbf{macroF1} \\
\hline
MAE-DFER & 0.509 & 0.208 & 0.179 & 0.222 & 0.206 & 0.339 & 0.696 & 0.237 & 0.347 & 0.569 &0.449 & 0.3602 \\ 
MAE-DFER FS & 0.341 & 0.225 & 0.192 & 0.083 & 0.100 & 0.411 & 0.645 & 0.163 & 0.332 & 0.536 & 0.318 & 0.3041 \\
\hline
HiCMAE & 0.611 & 0.296 & 0.269 & 0.164 & 0.464 & 0.448 & 0.689 & 0.042 & 0.410 & 0.535 & 0.466 & 0.3993 \\
HiCMAE FS & 0.392 & 0.235 & 0.200 & 0.118 & 0.197 & 0.471 & 0.700 & 0.043 & 0.349 & 0.603 & 0.371 & 0.3345 \\
\hline
Qwen2.5-VL & 0.354 & 0.102 & 0.103 & 0.070 & 0.196 & 0.409 & 0.539 & 0.000 & 0.247 & 0.455 & 0.219 & 0.2449 \\
Qwen2.5-VL FS & 0.351 & 0.073 & 0.170 & 0.066 & 0.245 & 0.382 & 0.500 & 0.083 & 0.240 & 0.413 & 0.233 & 0.2506 \\
\hline
Qwen2.5-Omni & 0.540 & 0.066 & 0.192 & 0.125 & 0.185 & 0.480 & 0.522 & 0.222 & 0.277 & 0.385 & 0.373 & 0.306 \\
Qwen2.5-Omni FS & 0.536 & 0.125 & 0.189 & 0.163 & 0.164 & 0.430 & 0.500 & 0.197 & 0.283 & 0.391 & 0.291 & 0.2972 \\
\hline
Qwen3-VL & 0.389 & 0.197 & 0.044 & 0.097 & 0.182 & 0.449 & 0.576 & 0.043 & 0.288 & 0.409 & 0.340 & 0.2738 \\
Qwen3-VL FS & 0.274 & 0.188 & 0.082 & 0.000 & 0.143 & 0.469 & 0.546 & 0.078 & 0.313 & 0.434 & 0.352 & 0.2615 \\
\hline
EmotionQwen & 0.348 & 0.053 & 0.219 & 0.000 & 0.178 & 0.259 & 0.641 & 0.041 & 0.303 & 0.430 & 0.367 & 0.2581 \\ 
EmotionQwen FS & 0.356 & 0.000 & 0.123 & 0.051 & 0.242 & 0.118 & 0.625 & 0.044 & 0.328 & 0.479 & 0.404 & 0.2517 \\ 
\hline
Video-LLaVA & 0.210 & 0.000 & 0.000 & 0.000 & 0.120 & 0.000 & 0.396 & 0.000 & 0.000 & 0.231 & 0.000 & 0.0870 \\
Video-LLaVA FS & 0.193 & 0.000 & 0.000 & 0.000 & 0.041 & 0.000 & 0.447 & 0.000 & 0.000 & 0.215 & 0.000 & 0.0814 \\
\hline
LLaVA-NeXT-V & 0.352 & 0.194 & 0.000 & 0.000 & 0.176 & 0.000 & 0.492 & 0.000 & 0.000 & 0.326 & 0.042 & 0.1438 \\
LLaVA-NeXT-V FS & 0.367 & 0.128 & 0.000 & 0.000 & 0.204 & 0.000 & 0.433 & 0.000 & 0.000 & 0.311 & 0.077 & 0.1382 \\
\hline
InternVL-3.0 & 0.270 & 0.129 & 0.143 & 0.000 & 0.234 & 0.381 & 0.469 & 0.080 & 0.240 & 0.424 & 0.320 & 0.2445 \\
InternVL-3.0 FS &  0.314 & 0.138 & 0.077 & 0.000 & 0.187 & 0.313 & 0.469 & 0.151 & 0.177 & 0.442 & 0.291 & 0.2326 \\
\hline
Gemini2.5-Flash & 0.537 & 0.263 & 0.333 & 0.262 & 0.213 & 0.587 & 0.600 & 0.091 & 0.273 & 0.532 & 0.444 & 0.3758 \\
Gemini2.5-Flash FS & 0.543 & 0.214 & 0.316 & 0.233 & 0.133 & 0.558 & 0.604 & 0.064 & 0.300 & 0.539 & 0.485 & 0.3626 \\
\hline
\end{tabular}}
\caption{Extended Analysis of Temporal Sensitivity on MAFW dataset. We report Per-Class and Macro-Averaged F1 scores for a wide range of architectures on Normal vs. Frame Shuffled (FS) videos. AG: Anger, AX: Anxiety, CO: Contempt, DA: Disappointment, DG: Disgust, FE: Fear, HA: Happiness, HP: Helplessness, NU: Neutral, SD: Sadness, SU: Surprise.}
\label{tab:supp_frame-shuffling-full-class}
\end{table*}

As detailed in Table~\ref{tab:supp_frame-shuffling-full-class} and~\ref{tab:supp_frame-shuffling-full-class-dfew}, the pattern holds: VLMs exhibit negligible performance drops (and often improvements) when frames are shuffled (FS). 
For instance, Qwen2.5-VL improved slightly with shuffling on the MAFW dataset (macro-F1 increasing from 0.2449 to 0.2506) and deteriorated only negligibly on the DFEW dataset (0.4552 to 0.4534). Similarly, Qwen3-VL demonstrated a slight performance improvement on DFEW (0.5511 to 0.5538) when temporal order was destroyed.
This provides definitive evidence that the model treats video frames as an order-agnostic `bag of frames'. Video-LLaVA, LLaVA-NeXT-Video, EmotionQwen, InternVL-3.0 and Gemini2.5-Flash showed minor drops, confirming that even video-specific tuning does not fundamentally alter the underlying aggregation mechanism to be temporally causal. Qwen2.5-Omni also shows a slight performance drop, signifying that adding audio cannot naively improve temporal understanding. This contrasts sharply with vision-only baselines (MAE-DFER, HiCMAE), which suffer massive drops when temporal order is destroyed.

\begin{table*}[htbp]
\centering
\small
\setlength{\tabcolsep}{4pt}
\begin{tabular}{l | c c c c c c c | c}
\hline
\textbf{Model} &
\textbf{AG} &
\textbf{DG} &
\textbf{FE} &
\textbf{HA} &
\textbf{NU} &
\textbf{SD} &
\textbf{SU} &
\textbf{macroF1} \\
\hline
MAE-DFER & 0.722 & 0.058 & 0.416 & 0.936 & 0.523 & 0.667 & 0.630 & 0.5645 \\ 
MAE-DFER FS & 0.563 & 0.000 & 0.484 & 0.847 & 0.456 & 0.650 & 0.362 & 0.4802 \\
\hline
HiCMAE & 0.762 & 0.020 & 0.497 & 0.908 & 0.562 & 0.687 & 0.571 & 0.5725 \\
HiCMAE FS & 0.604 & 0.000 & 0.465 & 0.843 & 0.504 & 0.695 & 0.248 & 0.4797 \\
\hline
Qwen2.5-VL & 0.585 & 0.196 & 0.447 & 0.678 & 0.401 & 0.447 & 0.432 & 0.4552 \\
Qwen2.5-VL FS & 0.599 & 0.156 & 0.443 & 0.709 & 0.401 & 0.453 & 0.413 & 0.4534 \\
\hline
Qwen2.5-Omni & 0.616 & 0.075 & 0.438 & 0.682 & 0.369 & 0.509 & 0.318 & 0.4296 \\
Qwen2.5-Omni FS & 0.604 & 0.073 & 0.456 & 0.626 & 0.355 & 0.525 & 0.3200 & 0.4226 \\
\hline
Qwen3-VL & 0.688 & 0.076 & 0.600 & 0.853 & 0.489 & 0.586 & 0.565 & 0.5511 \\
Qwen3-VL FS & 0.670 & 0.076 & 0.630 & 0.847 & 0.494 & 0.616 & 0.543 & 0.5538 \\
\hline
EmotionQwen & 0.632 & 0.073 & 0.397 & 0.809 & 0.473 & 0.614 & 0.509 & 0.5010 \\ 
EmotionQwen FS & 0.597 & 0.260 & 0.440 & 0.723 & 0.395 & 0.508 & 0.503 & 0.4895 \\ 
\hline
Video-LLaVA & 0.317 & 0.000 & 0.000 & 0.457 & 0.019 & 0.365 & 0.000 & 0.1654 \\
Video-LLaVA FS & 0.260 & 0.000 & 0.000 & 0.420 & 0.055 & 0.338 & 0.000 & 0.1531 \\
\hline
LLaVA-NeXT-V & 0.444 & 0.021 & 0.099 & 0.898 & 0.194 & 0.327 & 0.094 & 0.2969 \\
LLaVA-NeXT-V FS & 0.400 & 0.000 & 0.058 & 0.862 & 0.197 & 0.326 & 0.056 & 0.2712 \\
\hline
InternVL-3.0 & 0.634 & 0.118 & 0.364 & 0.845 & 0.435 & 0.609 & 0.526 & 0.5044 \\
InternVL-3.0 FS & 0.598 & 0.108 & 0.457 & 0.773 & 0.455 & 0.612 & 0.487 & 0.4985 \\
\hline
Gemini2.5-Flash & 0.754 & 0.143 & 0.600 & 0.881 & 0.507 & 0.695 & 0.626 & 0.6008 \\
Gemini2.5-Flash FS & 0.726 & 0.127 & 0.546 & 0.854 & 0.527 & 0.676 & 0.589 & 0.5778 \\
\hline
\end{tabular}
\caption{Extended Analysis of Temporal Sensitivity on DFEW dataset. We report Per-Class and Macro-Averaged F1 scores for a wide range of architectures on Normal vs. Frame Shuffled (FS) videos. AG: Angry, DG: Disgust, FE: Fear, HA: Happy, NU: Neutral, SD: Sad, SU: Surprise.}
\label{tab:supp_frame-shuffling-full-class-dfew}
\end{table*}

\noindent
\textit{Performance Anomaly in Video-LLaVA}: We attribute the exceptionally low performance of Video-LLaVA (F1: 0.0870 on MAFW dataset, and F1: 0.1654 on DFEW dataset) to its architectural bias towards coarse-grained action recognition. Unlike the other models, which process high-resolution frame tokens, Video-LLaVA employs aggressive token compression to align video features with the LLM. While efficient for describing distinct events (\eg, `running'), this compression discards the fine-grained high-frequency spatial details required to detect micro-expressions, resulting in a mode collapse where the model fails to predict any nuanced tail-class emotions (\eg, Contempt, Anxiety) entirely.

\section{Generalizability of `Attentional Dilution' (FPS Variation)}
\label{supsec:fps_variation}
\begin{figure}[htbp]
\centering
\includegraphics[width=0.7\linewidth]{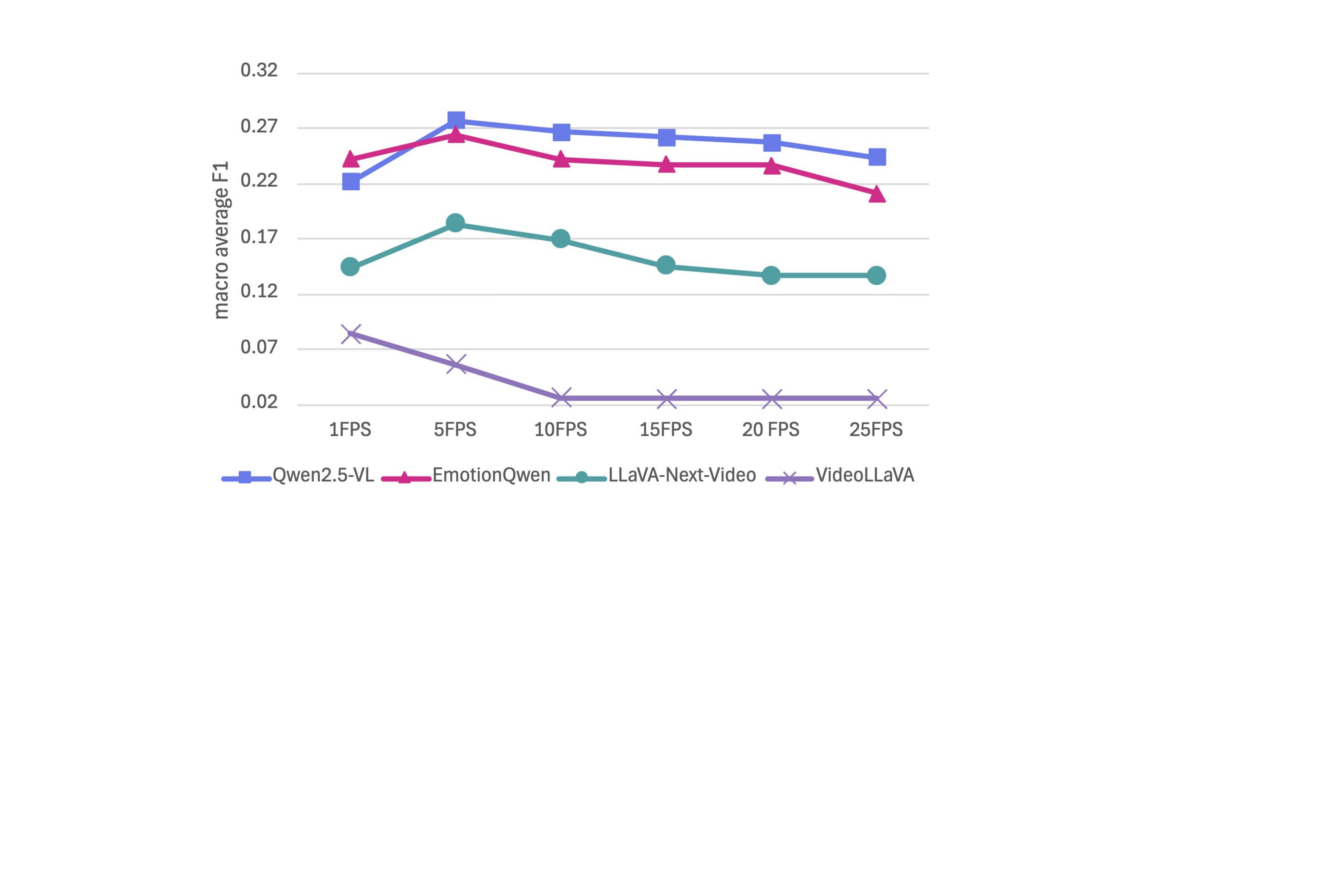}
  \caption{Plot of F1 Score vs. Frame Rate (FPS) for additional models (Video-LLaVA, LLaVA-NeXT-Video). All models exhibit similar trends: performance peaks at moderate sampling rates (5 FPS) and degradation at high frame rates (15-25 FPS). This confirms that increasing visual token density overwhelms the context window, leading to attentional dilution rather than better temporal understanding.}
  \label{fig:supp_fps}
\end{figure}

To verify that the performance curve reported in Figure~\ref{fig:fps-variation} (main paper), is not an artifact of the Qwen architecture, we replicated the FPS experiment on Video-LLaVA and LLaVA-NeXT-Video (Figure~\ref{fig:supp_fps}). 

All tested models exhibit similar characteristics.
Performance peaks at a moderate sampling rate (5 FPS) and significant degradation at higher frame rates (15-25 FPS). This provides additional support to the idea that redundant visual tokens overwhelm the context window and lead to attentional dilution. It shows a fundamental architectural bottleneck common to current transformer-based VLMs, regardless of the specific pre-training recipe. For Video-LLaVA, we attribute the sharp performance drop due to very small context window, which collapses even with 10FPS.

\section{Data Distribution Analysis on Vision-Only Models}
\label{supsec:non-vlms}

\begin{figure}[htbp]
\centering
\includegraphics[width=\linewidth]{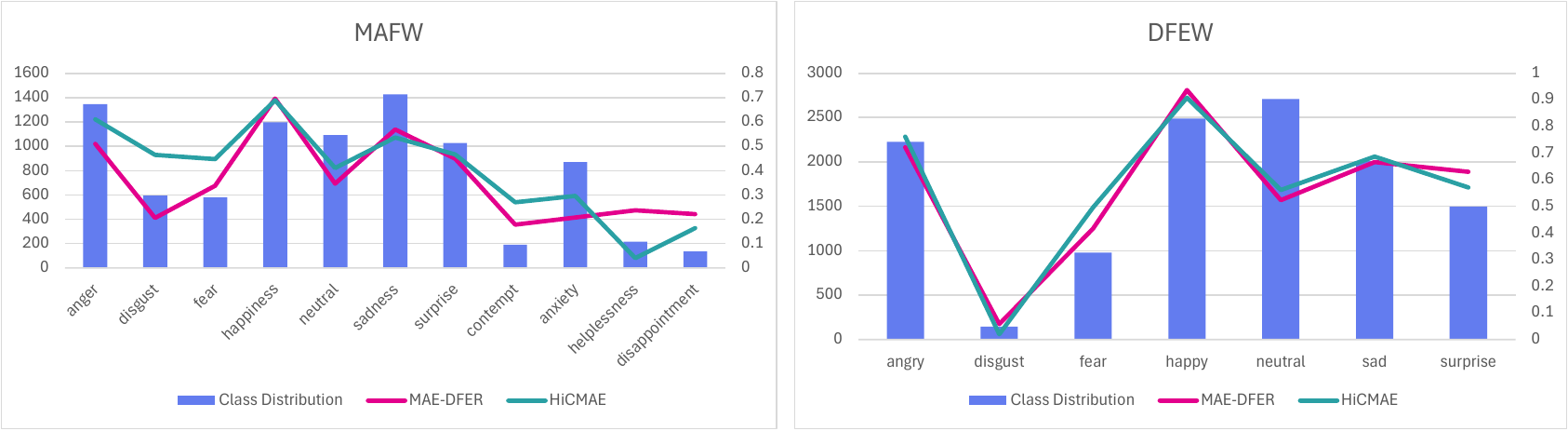}
  \caption{
  Overlay of per-class F1 scores for vision-only classifiers (MAE-DFER and HiCMAE) with the class frequency distributions of the MAFW and DFEW datasets. A strong correlation between class frequency and model performance is observed: `Head' classes (e.g., Happiness and Sadness in MAFW; Happy and Sad in DFEW) achieve high F1 scores, while `Tail' classes (e.g., Contempt and Helplessness in MAFW; Disgust and Fear in DFEW) are severely under-predicted, mirroring the class-imbalance bias commonly observed in VLMs.}
  \label{fig:supp_data}
\end{figure}

To further validate that the long-tail performance drop is a data-driven issue (rather than just a VLM-specific flaw), we visualize the performance of vision-only classifiers (MAE-DFER, HiCMAE) against the dataset distribution of MAFW and DFEW. Figure~\ref{fig:supp_data} illustrates the per-class F1 scores overlaid on the class frequency distribution. We observe a strong correlation: `Head' classes (\eg, Happiness, Sadness) achieve high F1 scores ($>$ 0.5), `Tail' classes (\eg, Contempt, Helplessness) drop precipitously ($<$0.2) in MAFW dataset, mirroring the behavior we observed in VLMs. We see a similar pattern in DFEW dataset, where `Head' classes (\eg, Happy, Sad) achieve high F1 scores ($>$ 0.7), `Tail' classes (\eg, Disgust, Fear) drop precipitously ($<$0.3). This supports our hypothesis that the `emotion understanding gap' is partially rooted in the fundamental imbalance of affective data available for training, affecting both specialized vision models and generalist VLMs alike.

\section{Directional Asymmetry of Confusion}
\label{sec:supp-dai}
In Sec.~\ref{subsec:long-tail-data-distribution} we noted that errors are
asymmetric: rare classes collapse into common ones but not the reverse. Here
we formalise this beyond visual inspection of the confusion matrices. For a
rare class $r$ and a common class $c$, let $f(r{\to}c)$ denote the fraction of
test videos with ground-truth label $r$ that the model predicts as $c$, i.e.\
a row-normalised entry of the confusion matrix (Table~\ref{tab:supp-confmat}).
We define a \emph{Directional Asymmetry Index}
\begin{equation}
\Delta(r, c) = f(r{\to}c) - f(c{\to}r),
\label{eq:dai}
\end{equation}
which is positive when a rare class $r$ collapses into a common class $c$ more
often than the reverse, and near zero when the two classes are confused
symmetrically. The index is antisymmetric ($\Delta(r,c) = -\Delta(c,r)$), so a
value near zero is the expected signature of pure semantic ambiguity.

Table~\ref{tab:supp-confmat} reports the underlying Qwen2.5-VL confusion matrix on the MAFW balanced test split (45 videos per class), and Table~\ref{tab:supp-dai} reports $\Delta$ for representative pairs. 
The rare$\to$common collapses are strongly directional: low-frequency 
emotions (contempt, disappointment, helplessness, anxiety) are absorbed into the dominant \emph{neutral} class with $\Delta$ up to $+0.556$, while the reverse transition almost never occurs. In contrast, a semantically adjacent head-class pair such as Anger/Disgust is near-symmetric ($|\Delta|=0.022$).
This directional gap supports our claim that errors are driven by collapse toward high-frequency priors rather than by bidirectional semantic confusion.

\begin{table}[htbp]
\vspace{-0.5cm}
\centering
\small
\setlength{\tabcolsep}{4pt}
\begin{tabular}{l ccccccccccc}
\toprule
True\,$\downarrow$ / Pred\,$\rightarrow$ & AG & AX & CO & DA & DG & FE & HA & HP & NU & SD & SU \\
\midrule
Anger (AG)          & \textbf{14} & 2 & 1 & 1 & 8 & 2 & 3 & 0 & 9 & 4 & 1 \\
Anxiety (AX)        & 2 & \textbf{3} & 3 & 2 & 6 & 4 & 0 & 1 & 19 & 4 & 1 \\
Contempt (CO)       & 3 & 0 & \textbf{3} & 1 & 5 & 1 & 1 & 0 & 26 & 4 & 1 \\
Disappointment (DA) & 0 & 0 & 0 & \textbf{2} & 4 & 0 & 0 & 0 & 26 & 12 & 1 \\
Disgust (DG)        & 7 & 0 & 1 & 1 & \textbf{10} & 1 & 2 & 0 & 16 & 5 & 2 \\
Fear (FE)           & 3 & 1 & 2 & 1 & 6 & \textbf{18} & 2 & 0 & 1 & 6 & 5 \\
Happiness (HA)      & 1 & 0 & 0 & 0 & 2 & 2 & \textbf{21} & 0 & 18 & 1 & 0 \\
Helplessness (HP)   & 1 & 2 & 2 & 1 & 1 & 3 & 1 & \textbf{0} & 25 & 9 & 0 \\
Neutral (NU)        & 1 & 2 & 1 & 1 & 3 & 3 & 1 & 0 & \textbf{29} & 4 & 0 \\
Sadness (SD)        & 1 & 1 & 0 & 2 & 2 & 1 & 0 & 1 & 8 & \textbf{28} & 1 \\
Surprise (SU)       & 1 & 3 & 0 & 0 & 10 & 8 & 2 & 0 & 13 & 1 & \textbf{7} \\
\bottomrule
\end{tabular}
\caption{Qwen2.5-VL confusion matrix on the MAFW balanced test split (45 videos per class). Rows are ground-truth labels, columns are predictions. Column abbreviations follow Table~S2: AG Anger, AX Anxiety, CO Contempt, DA Disappointment, DG Disgust, FE Fear, HA Happiness, HP Helplessness, NU Neutral, SD Sadness, SU Surprise. Note the dense column under \textbf{NU}, into which rare classes collapse.}
\label{tab:supp-confmat}
\vspace{-1.0cm}
\end{table}

\begin{table}[htbp]
\vspace{-1.0cm}
\centering
\small
\setlength{\tabcolsep}{6pt}
\begin{tabular}{ll ccc}
\toprule
\multicolumn{2}{l}{Pair $(r \to c)$} & $f(r{\to}c)$ & $f(c{\to}r)$ & $\Delta(r,c)$ \\
\midrule
\multicolumn{5}{l}{\emph{Rare $\to$ common (collapse)}} \\
Helplessness   & $\to$ Neutral & 0.556 & 0.000 & $+0.556$ \\
Contempt       & $\to$ Neutral & 0.578 & 0.022 & $+0.556$ \\
Disappointment & $\to$ Neutral & 0.578 & 0.022 & $+0.556$ \\
Anxiety        & $\to$ Neutral & 0.422 & 0.044 & $+0.378$ \\
Disgust        & $\to$ Neutral & 0.356 & 0.067 & $+0.289$ \\
Disappointment & $\to$ Sadness & 0.267 & 0.044 & $+0.222$ \\
Helplessness   & $\to$ Sadness & 0.200 & 0.022 & $+0.178$ \\
\midrule
\multicolumn{5}{l}{\emph{Semantically adjacent head-class controls}} \\
Anger & $\leftrightarrow$ Disgust  & 0.178 & 0.156 & $+0.022$ \\
Anger & $\leftrightarrow$ Sadness  & 0.089 & 0.022 & $+0.067$ \\
Fear  & $\leftrightarrow$ Surprise & 0.111 & 0.178 & $-0.067$ \\
\bottomrule
\end{tabular}
\caption{Directional Asymmetry Index $\Delta(r,c) = f(r{\to}c) - f(c{\to}r)$ for representative class pairs, computed from Table~\ref{tab:supp-confmat}. \emph{Top:} rare$\to$common pairs exhibit large positive asymmetry, indicating systematic collapse of low-frequency emotions into high-frequency ones. \emph{Bottom:} semantically adjacent head-class control pairs are near-symmetric ($|\Delta|$ small), as expected if confusion were purely semantic.}
\label{tab:supp-dai}
\vspace{-1.0cm}
\end{table}

\section{Per-Class Analysis of Balanced Fine-Tuning}
\label{sec:supp-longtail}
To complement Sec.~\ref{subsec:mitigate-long-tail}, we report the full per-class Precision, Recall, and F1 for Qwen2.5-VL on the MAFW balanced test split, both zero-shot and after LoRA balanced fine-tuning (Table~\ref{tab:supp-loraft}). In the zero-shot setting, predictions are dominated by the head class: $190$ of $495$ predictions fall on \emph{Neutral}, which attains high recall ($0.64$) but low precision ($0.15$) as rare classes collapse into it, while Helplessness is never predicted correctly (F1 $=0.00$).
Balanced fine-tuning raises macro-F1 from $0.245$ to $0.315$ and yields more uniform prediction distribution: the dominant prediction count drops from $190$ to $117$, and $8$ of $11$ classes improve in F1.
The largest gains are on tail classes the zero-shot model effectively ignored, notably Helplessness ($0.00\!\to\!0.28$) and Disgust ($0.20\!\to\!0.28$), confirming that under-performance on rare emotions reflects a data-driven bias rather than an inherent incapacity.
Consistent with rebalancing \emph{redistributing} rather than eliminating the bias, error mass shifts away from Neutral and onto Helplessness, which becomes over-predicted (recall $0.51$, precision $0.20$).
Disappointment is the one tail class that does not benefit, indicating it remains genuinely difficult to disambiguate from adjacent classes such as Sadness.

\begin{table}[htbp]
\centering
\small
\setlength{\tabcolsep}{6pt}
\begin{tabular}{l ccc ccc}
\toprule
& \multicolumn{3}{c}{Zero-shot} & \multicolumn{3}{c}{Balanced FT} \\
\cmidrule(lr){2-4}\cmidrule(lr){5-7}
Class & P & R & F1 & P & R & F1 \\
\midrule
Anger          & 0.412 & 0.311 & 0.354 & 0.312 & 0.422 & 0.359 \\
Anxiety        & 0.214 & 0.067 & 0.102 & 0.250 & 0.156 & 0.192 \\
Contempt       & 0.231 & 0.067 & 0.103 & 0.353 & 0.133 & 0.194 \\
Disappointment & 0.167 & 0.044 & 0.070 & 0.250 & 0.022 & 0.041 \\
Disgust        & 0.175 & 0.222 & 0.196 & 0.314 & 0.244 & 0.275 \\
Fear           & 0.419 & 0.400 & 0.409 & 0.500 & 0.378 & 0.430 \\
Happiness      & 0.636 & 0.467 & 0.539 & 0.560 & 0.622 & 0.590 \\
Helplessness   & 0.000 & 0.000 & 0.000 & 0.197 & 0.511 & 0.284 \\
Neutral        & 0.153 & 0.644 & 0.247 & 0.254 & 0.333 & 0.288 \\
Sadness        & 0.359 & 0.622 & 0.455 & 0.460 & 0.511 & 0.484 \\
Surprise       & 0.368 & 0.156 & 0.219 & 0.350 & 0.311 & 0.329 \\
\midrule
Macro avg      & 0.285 & 0.273 & 0.245 & 0.345 & 0.331 & 0.315 \\
\bottomrule
\end{tabular}
\caption{Per-class Precision (P), Recall (R), and F1 for Qwen2.5-VL on the MAFW balanced test split ($45$ videos per class): zero-shot vs.\ LoRA balanced fine-tuning. Balanced fine-tuning raises macro-F1 from $0.245$ to $0.315$ and lifts most tail classes (e.g.\ Helplessness $0.00\!\to\!0.28$, Disgust $0.20\!\to\!0.28$). Consistent with rebalancing redistributing rather than eliminating the head-class sink, error mass shifts from Neutral toward Helplessness; Disappointment is the one tail class that does not improve.}
\label{tab:supp-loraft}
\end{table}

\end{document}